\documentclass[sigconf]{acmart}

\usepackage{times}
\usepackage{latexsym}
\usepackage{algorithm}
\usepackage{algorithmic}
\usepackage{amsmath}
\usepackage{tabularx}   
\usepackage{booktabs}
\usepackage{amssymb}
\usepackage{microtype}
\usepackage{tikz}
\usepackage{tcolorbox}
\usepackage{amsmath}

\usepackage{graphicx}
\usepackage{subcaption} 
\usepackage{caption} 
\usepackage{float} 
\AtBeginDocument{%
  }

\author{YUKUN ZHANG}
\affiliation{%
  \institution{The Chinese University Of Hongkong}
  \city{HongKong}
  \country{China}}
\email{215010026@link.cuhk.edu.cn}

    \author{QI DONG}
\affiliation{%
  \institution{Fudan University}
  \city{ShangHai}
  \country{China}}
\email{19210980065@fudan.edu.cn}

\begin{document}

\title{Unveiling LLM Mechanisms Through Neural ODEs and Control Theory}

\begin{abstract}
  This paper proposes a framework combining Neural Ordinary Differential Equations (Neural ODEs) and robust control theory to enhance the interpretability and control of large language models (LLMs). By utilizing Neural ODEs to model the dynamic evolution of input-output relationships and introducing control mechanisms to optimize output quality, we demonstrate the effectiveness of this approach across multiple question-answer datasets. Experimental results show that the integration of Neural ODEs and control theory significantly improves output consistency and model interpretability, advancing the development of explainable AI technologies.

\end{abstract}




\maketitle

\section{Introduction}
Large Language Models (LLMs) have demonstrated impressive performance across a range of natural language processing tasks, from machine translation to text generation and summarization\cite{brown2020languagemodelsfewshotlearners}. Despite their remarkable capabilities,interpreting the decision-making processing of these models remains a significant challenge. The opacity of LLMs raises critical questions about their reliability, fairness, and ethical implication in real-word application \cite{lipton2017mythosmodelinterpretability}. Understanding the underlying mechanisms of LLMs is essential for building trust and accountability in AI systems, particularly when these systems are deployed in high-stakes environments such as healthcare and low.

\section{Literature Review}

\subsection{Current Methods for Enhancing Interpretability in LLMs}
The interpretability of large language models (LLMs) has become a central concern in AI research. To address this challenge, various approaches have been proposed, which can broadly be categorized into local and global analyses.

\subsubsection{Local Analysis}
Local analysis focuses on interpreting individual predictions made by a model by examining specific input-output relationships. The primary goal is to understand the contribution of each input feature to the model's output. Key approaches within local analysis include feature attribution methods and analyzing transformer blocks.

Feature attribution methods quantify the influence of each input feature on the model’s predictions. Common techniques include gradient-based methods and vector-based methods. Gradient-based methods, such as Integrated Gradients \cite{sundararajan2017axiomaticattributiondeepnetworks}, compute the gradients of the output with respect to inputs, attributing significance based on how changes in input features affect the model’s predictions. Vector-based methods, such as the Shapley Value framework \cite{lundberg2017unifiedapproachinterpretingmodel}, evaluate the contribution of each feature by considering all possible combinations of input variables. These methods allow researchers to identify influential features and gain insights into why the model produces specific outputs.

For transformer-based models like BERT and GPT, analyzing components such as multi-head self-attention (MHSA) and multi-layer perceptron (MLP) sublayers can provide valuable insights into the model’s behavior. For example, examining MHSA sublayers reveals how attention weights are distributed across input tokens, helping to determine if the model focuses on relevant words or phrases \cite{vig2019visualizingattentiontransformerbasedlanguage}. Similarly, analyzing MLP sublayers reveals how feature combinations are processed and transformed, elucidating the flow of information through the network. These analyses can uncover how specific tokens drive shifts in attention across layers, leading to richer interpretations of model decisions \cite{beltagy2020longformerlongdocumenttransformer}.

\subsubsection{Global Analysis}
In contrast to local analysis, global analysis seeks to provide a broader understanding of a model’s behavior by exploring the underlying principles that govern its representations and knowledge. This approach includes probing-based methods and mechanistic interpretability.

Probing-based methods involve training auxiliary models on hidden representations to assess the knowledge encoded within the model. Techniques such as probing knowledge and probing representations allow researchers to evaluate the extent to which LLMs capture linguistic properties, syntactic structures, or semantic features \cite{tenney2019learncontextprobingsentence}. These methods help identify the high-level knowledge embedded in the model, offering insights into its internal decision-making processes.

Mechanistic interpretability focuses on understanding how specific mechanisms within the model contribute to its predictions. Techniques like causal tracing \cite{meng2023locatingeditingfactualassociations} provide insights into how neural networks, especially LLMs like GPT, represent and utilize factual knowledge. By locating factual associations, researchers can identify which parts of the model are responsible for generating specific factual outputs.

While local analysis provides insights into specific model decisions, global analysis offers a more holistic view of the model's overall behavior. These two approaches are complementary: local analysis helps interpret individual predictions, while global analysis aids in understanding how the model generalizes across different tasks and domains.

\subsection{Neural ODEs in LLMs}
Neural Ordinary Differential Equations (ODEs) have emerged as a powerful framework for continuous-time modeling, with applications across various domains, including language models (LLMs).

Introduced by Chen et al. (2019) \cite{chen2019neuralordinarydifferentialequations}, Neural ODEs model the evolution of latent variables as a continuous-time dynamical system. Unlike traditional neural networks, which rely on discrete layers, Neural ODEs define a differential equation parameterized by neural networks, allowing for flexible representations of data evolving over time. Neural ODEs are particularly useful for modeling the temporal patterns seen in language processing, enabling the model to adaptively learn how different linguistic structures evolve \cite{rubanova2019latentodesirregularlysampledtime}.

Additionally, integrating Neural ODEs with attention mechanisms has led to scalable LLMs capable of processing and interpreting real-time sensor data \cite{wang2024learningadaptivehydrodynamicmodels}. A promising direction is combining Neural ODEs with Transformer architectures, which has revealed natural correspondences between Neural ODEs and Transformer attention mechanisms, offering new insights into deep learning models \cite{hashimoto2024unificationsymmetriesinsideneural}.

However, despite their potential, current applications of Neural ODEs in understanding neural models often fall short in directly correlating these dynamics with specific input-output relationships, particularly in complex architectures like LLMs. Existing frameworks typically do not address how these learned dynamics can be tuned based on external objectives, such as ensuring fairness or robustness. This presents a significant challenge in applying Neural ODEs to real-world LLM applications.

\subsection{Control Theory in LLMs}
Control theory offers critical insights into the dynamics and optimization of complex systems, making it highly relevant for improving the interpretability and reliability of Large Language Models (LLMs). The application of robust control helps address uncertainty within these models and ensures they meet performance standards.

Control is crucial in LLM research. As noted by \cite{liang2024controllabletextgenerationlarge}, controllable text generation aims to generate text according to specific requirements, including content and attribute control. Fine-tuning and retraining adjust the model during training, while reinforcement learning and prompt engineering guide the model during inference to enhance text controllability. Researchers in control engineering \cite{kevian2024capabilitieslargelanguagemodels} use specialized datasets and evaluation methods to understand the problem-solving capabilities of different advanced LLMs in control engineering contexts, guiding future improvements.

Control theory in LLMs is an evolving field. By exploring text generation, safety, multimodal tasks, and domain-specific applications, control theory can make LLMs more interpretable, reliable, and powerful, with broad applications in areas such as autonomous systems, healthcare, and finance.

\section{Contributions and Structure of the Paper}

This paper introduces an innovative approach for enhancing the interpretability and reliability of Large Language Models (LLMs) by integrating Neural Ordinary Differential Equations (Neural ODEs) with robust control mechanisms. The key contributions are as follows:

\begin{itemize}
    \item \textbf{Innovative Integration Method}: We pioneer the integration of Neural ODEs and robust control for LLMs, offering a new perspective on analyzing input-output relationships in LLMs that has not been explored before.
    \item \textbf{Enhanced Model Understanding}: This work enables the transformation of LLM inputs and outputs into a lower-dimensional latent space, providing a means to study internal information-processing pathways, which significantly enhances model interpretability.
    \item \textbf{Improved Output Quality}: The application of robust control ensures that LLM outputs meet specific performance criteria, improving their quality and reliability for practical use.
\end{itemize}

The paper is structured as follows: Section 4 utilizes Ordinary Differential Equations (ODEs) to model LLM processes, offering a continuous and interpretable framework, with robust control mechanisms introduced to enhance output reliability and ensure ethical standards. Section 5 presents the methodological framework, integrating Neural ODEs with and without control mechanisms to model dynamic processes within LLMs. Section 6 provides a comparative analysis of Neural ODE models, with and without control, focusing on training/validation loss, prediction accuracy, and latent space dynamics. Finally, Section 7 demonstrates how integrating control mechanisms into Neural ODEs enhances LLM stability and generalization, which is crucial for developing trustworthy AI in high-stakes domains, setting the stage for future research on transparent and accountable AI technologies.

\section{Theoretical Framework}

\begin{figure*}[htbp]
    \centering
    \includegraphics[width=\linewidth]{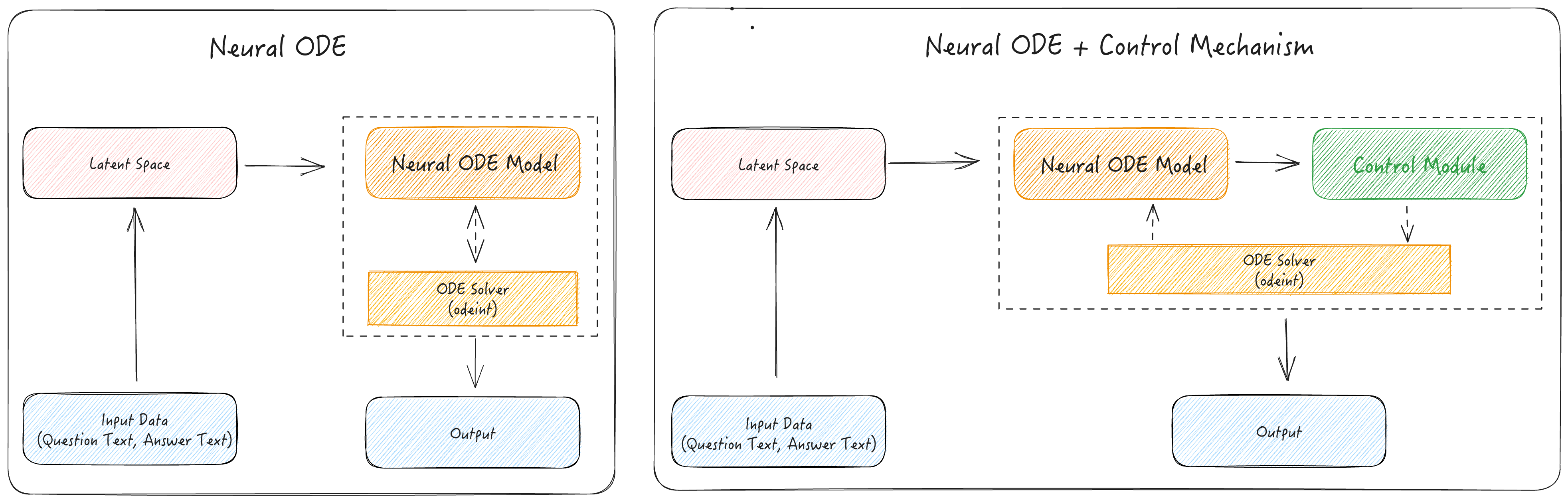}
    \caption{Model Architecture for Methodology}
    \label{fig:architecture_comparison}
\end{figure*}

\subsection{Neural ODEs}

Because Ordinary Differential Equations (ODEs) are inherently unable to model text directly, we need to map the input and output of Large Language Models (LLMs) into a latent space. This latent space representation allows us to work with continuous-time dynamics, enabling the use of Neural ODEs for modeling the evolution of LLM inputs and outputs in a more flexible and accurate manner.
Neural Ordinary Differential Equations (Neural ODEs) offer a powerful framework for modeling continuous-time dynamics. Their application in large language models (LLMs) provides a way to better capture the temporal relationships inherent in language data. Unlike traditional discrete models (e.g., RNNs or LSTMs), which treat sequences of data as a series of steps, Neural ODEs model the evolution of latent variables continuously over time, offering a more flexible and accurate representation of sequential data.

In the context of LLMs, we define the hidden state \( h(t) \) as a function of time \( t \). The evolution of this state is described by the following equation:

\begin{equation}
     \frac{dh(t)}{dt} = f(t, h(t), \theta)
\end{equation}

where \( f \) is a function parameterized by a neural network. For simplicity, assume that \( f \) is a single-layer neural network with a linear transformation followed by an activation function \( \sigma \):

\begin{equation}
     f(t, h(t), \theta) = \sigma(W \cdot h(t) + b)
\end{equation}

Here, \( W \) is the weight matrix, \( b \) is the bias vector, and \( \theta \) represents the parameters of the network.

This continuous modeling approach is particularly useful for time-series and language processing tasks, where data evolves over time. To solve the differential equation, numerical methods like Euler’s method can be applied:

\begin{equation}
     h(t + \Delta t) \approx h(t) + \Delta t \cdot f(t, h(t), \theta)
\end{equation}

This update rule models how the hidden state evolves over small time steps, enabling the model to learn how language structures change dynamically. Compared to traditional discrete models like RNNs or LSTMs, Neural ODEs offer several advantages, including more natural modeling of continuous temporal patterns and flexibility in capturing complex dependencies over time.

The mapping of LLM inputs and outputs to a latent space allows us to model the input-output relationships using continuous-time dynamics. Specifically, the input sequence \( \mathbf{X} = [x_1, x_2, \dots, x_T] \) is first embedded into a latent space \( \mathcal{Z} \) via an embedding function \( \phi \):

\begin{equation}
    \mathcal{Z}_t = \phi(\mathbf{X}_t)
\end{equation}

where \( \mathcal{Z}_t \) represents the embedded representation of the input at time step \( t \), and \( \phi \) is the function mapping the raw input tokens into the latent space. The output sequence \( \mathbf{Y} = [y_1, y_2, \dots, y_T] \) is then modeled in the same latent space, allowing the Neural ODE to capture the temporal evolution and dynamic mapping between the inputs and the corresponding outputs.

This formulation enables Neural ODEs to model the continuous-time evolution of LLM states, offering a powerful tool for handling sequential data with complex dependencies, such as language. By operating in the latent space, the Neural ODE framework is able to handle the high-dimensional, variable-length sequences typical of language models more efficiently and flexibly.

\subsection{Control Mechanism}
To improve the reliability of the LLM outputs, we introduce a robust control mechanism. The goal is to minimize the difference between the model's output \( y \) and the desired output \( y_{desired} \). The cost function \( J \) is defined as:

\begin{equation}
     J = \sum_{i = 1}^{n} w_i \cdot L_i(y, y_{desired})
\end{equation}

where \( L_i \) represents the individual loss components, and \( w_i \) are their respective weights. This cost function quantifies the discrepancy between the predicted and desired outputs across different loss components.

The output \( y \) depends on the hidden state \( h \) and control input \( u \), and is given by:

\begin{equation}
     y = g(h, u)
\end{equation}

where \( g(h, u) = V \cdot (h + u) + c \) is a neural network-based function, with \( V \) and \( c \) as additional parameters. The control input \( u \) is adjusted during training to minimize the cost function \( J \). This can be formulated as:

\begin{equation}
    \begin{split}
        u^*(t) &= \arg\min_{u} J(u) \\
        &= \arg\min_{u} \sum_{i=1}^{n} w_i \cdot L_i(g(h, u), y_{desired})
    \end{split}
\end{equation}

By optimizing the control input \( u \), we ensure that the model’s output \( y \) closely aligns with the desired output. This control mechanism enhances the LLM’s stability and reliability, making it more robust in real-world applications.

\subsection{Integrating Neural ODE and Control Mechanism}
To further enhance the LLM’s performance, we integrate the Neural ODE with the control mechanism. By incorporating the control input \( u \) into the Neural ODE, the evolution of the hidden state \( h(t) \) is modeled as:

\begin{equation}
     \frac{dh(t)}{dt} = f(t, h(t), \theta, u)
\end{equation}

In this combined framework, the cost function \( J \) depends on the entire sequence of outputs \( \{y(t)\} \), where \( y(t) = g(h(t), u(t)) \) for \( t = 0, \Delta t, \dots, T \). The optimization problem is then formulated as:

\begin{equation}
   (u^*, h^*) = \arg\min_{u,h} J(\{g(h(t), u(t))\}_{t=0}^T)
\end{equation}

This integrated approach ensures that the LLM’s output reflects both the internal processing of input data and the external control criteria. By optimizing both the hidden state and the control input, we can improve the model’s stability, generalization, and adaptability to various tasks.

The combination of Neural ODEs and control mechanisms allows for a more flexible and interpretable model. This approach is particularly beneficial for high-stakes applications where the model needs to be both interpretable and reliable. Furthermore, the continuous-time nature of Neural ODEs, combined with the control theory, provides a robust framework for optimizing model performance across a wide range of tasks, including real-time language processing and decision-making in dynamic environments.

\subsection{Conclusion}
In this section, we introduced a theoretical framework that integrates Neural ODEs with control mechanisms to optimize the performance and interpretability of LLMs. This approach provides flexibility in modeling sequential data and ensures that the model output meets specific performance criteria. Additionally, exploring how this framework can be adapted to handle diverse types of language data, such as multimodal inputs or noisy real-world data, would significantly expand its applicability. Further investigations into the interpretability of the combined model could also help provide deeper insights into how both internal dynamics and control inputs contribute to the model’s decision-making process.

\section{Methodology}

Building on the theoretical foundations established in the previous section, this section presents two algorithmic frameworks designed to enhance the interpretability and reliability of large language models (LLMs). These frameworks integrate Neural ODEs (Ordinary Differential Equations) and robust control mechanisms. The first framework utilizes Neural ODEs alone, encoding LLM inputs and outputs into a latent space and evolving the state using advanced optimization techniques. The second framework incorporates control mechanisms into the Neural ODEs, allowing for dynamic adjustment of the model's state to achieve stable and reliable outputs. Both frameworks aim to improve the transparency and performance of LLMs by revealing their continuous and dynamic transformations.

\subsection{Neural ODE for LLM Input-Output Mapping}

The first algorithm uses a basic Neural ODE framework to model the input-output relationships in LLMs without any control mechanisms. This framework consists of three key components, detailed in Algorithm~\ref{alg:neural_ode_minimal}.

\begin{algorithm}[htbp]
\caption{Train Neural ODE for LLM Input-Output Mapping}
\label{alg:neural_ode_minimal}
\begin{algorithmic}[1]
    \STATE \textbf{Input:} Dataset $(Q, A)$, Parameters $\theta$, Learning rate $\alpha$, Epochs $E$
    \STATE \textbf{Output:} Optimized parameters $\theta^*$
    
    \STATE Initialize model $\mathcal{M} \gets \text{NeuralODE}(\theta)$
    \STATE Initialize optimizer $Opt \gets \text{Adam}(\mathcal{M}, \alpha)$
    
    \FOR{epoch $= 1$ to $E$}
        \FOR{each $(q, a) \in (Q, A)$}
            \STATE $h \gets \text{Integrate}(\mathcal{M}, q)$
            \STATE $loss \gets \text{MSE}(h, a)$
            \STATE $Opt.\text{step}(loss)$
        \ENDFOR
    \ENDFOR
    
    \STATE \textbf{Return:} $\theta^*$
\end{algorithmic}
\end{algorithm}

The  algorithm models the relationships between inputs and outputs using Neural ODEs without incorporating any additional control mechanisms. The architecture consists of three key components: the input layer, which transforms raw input tokens into a lower-dimensional latent space using an embedding layer; the Neural ODE block, which models the hidden state dynamics as a continuous-time evolution governed by the Neural ODE; and the output layer, which maps the final hidden state to the desired output using a fully connected (dense) layer followed by an activation function.

\subsection{Neural ODE with Control Mechanism}
The second algorithm incorporates a control mechanism into the Neural ODE framework. This framework is designed to dynamically adjust the model's hidden state and ensure output reliability. The key components of this model are outlined in Algorithm~\ref{alg:neural_ode_with_control_optimized}.

\begin{algorithm}[htbp]
\caption{Train Neural ODE with Control for LLMs}
\label{alg:neural_ode_with_control_optimized}
\begin{algorithmic}[1]
    \STATE \textbf{Input:} Dataset $(Q, A)$, Parameters $\theta$, Learning rate $\alpha$, Epochs $E$, Control type $c$
    \STATE \textbf{Output:} Optimized parameters $\theta^*$
    
    \STATE Initialize model $\mathcal{M} \gets \text{NeuralODE}(\theta)$
    \STATE Initialize optimizer $Opt \gets \text{Adam}(\mathcal{M}, \alpha)$
    
    \FOR{epoch $= 1$ to $E$}
        \FOR{each $(q, a) \in (Q \cup Q_{test}, A \cup A_{test})$}
            \STATE $h \gets q$ \COMMENT{Initialize hidden state}
            \FOR{t $= 1$ to $T$}
                \STATE $f \gets \text{Dynamics}(h, \theta)$ \COMMENT{Compute dynamics}
                \STATE $H \gets \text{Control}(h, q, c)$ \COMMENT{Apply control}
                \STATE $h \gets h + \Delta t \cdot (f + H)$ \COMMENT{Update hidden state}
            \ENDFOR
            \STATE $loss \gets \text{MSE}(h, a)$ \COMMENT{Compute loss}
            \STATE $Opt.\text{step}(loss)$ \COMMENT{Optimize model parameters}
        \ENDFOR
    \ENDFOR
    
    \STATE \textbf{Return:} $\theta^*$
\end{algorithmic}
\end{algorithm}

The algorithm integrates a robust control mechanism into the Neural ODE framework, dynamically adjusting the hidden state to improve output reliability. The architecture includes the input layer, which transforms raw input tokens into embeddings; the Neural ODE block with control, which models hidden state dynamics while incorporating a control input \( u(t) \) to guide the state evolution; the control module, which computes the optimal control input based on the current hidden state and predefined standards; and the output layer, which maps the controlled hidden state to the desired output.

\section{Experiment and Results}
The experiments aim to validate these two methods' contributions to improving model performance in various contexts. Specifically, we conduct two experiments:

Experiment I visualizes the input-output relationships using Neural ODEs across multiple question-answer (QA) datasets from diverse domains, demonstrating the ability of Neural ODEs to capture complex input-output dynamics. Experiment II, on the other hand, applies Control Theory to regulate LLM outputs, assessing the impact of control mechanisms on model performance and reliability, particularly in terms of consistency and stability.

\subsection{Experimental Data}

\paragraph{Experiment I}

For Experiment I, we selected six distinct QA datasets that cover a range of domains, including factual knowledge bases, commonsense reasoning tasks, medical information, mathematical problem-solving, and truthful response generation. These datasets allow us to assess the versatility and adaptability of Neural ODEs in diverse settings. Table~\ref{tab:experiment1_datasets} provides an overview of these datasets, detailing their repositories, sizes, and specific tasks.

\paragraph{Experiment II}

We utilized the \texttt{aligner/aligner-20K} dataset, which consists of 20,000 aligned QA pairs carefully curated to ensure high relevance and accuracy between the input questions and their corresponding answers. This dataset is ideal for assessing the role of Control Theory in stabilizing model outputs. Experiment II is designed to evaluate the effectiveness of Control Theory in regulating model outputs and improving consistency.

\subsection{Experiment I: Results and Analysis}

For Experiment I, the Neural ODE model was trained on six different QA datasets. The training losses across epochs were recorded to assess the model’s convergence behavior. Table~\ref{tab:training_validation_losses_experiment1} presents the training and validation losses for each dataset.

\begin{table}[htbp]
    \centering
    \caption{Training and Validation Losses at Epoch 30 for QA Datasets in Experiment I}
    \label{tab:training_validation_losses_experiment1}
    \begin{tabularx}{\linewidth}{Xp{2cm}p{2cm}}
        \toprule
        \textbf{Dataset Name} & \textbf{Training Loss} & \textbf{Validation Loss} \\
        \midrule
        commonsense\_qa & 0.0290 & 0.0610 \\
        GammaCorpus-fact-qa & 0.0278 & 0.0578 \\
        medical-qa & 0.0053 & 0.0224 \\
        rvv-karma\_Math-QA & 0.0048 & 0.0086 \\
        trivia\_qa& 0.0291 & 0.0597 \\
        TruthfulQA & 0.0052 & 0.0339 \\
        \bottomrule
    \end{tabularx}
\end{table}

\begin{figure}[htbp]
    \centering
    \includegraphics[width=\columnwidth]{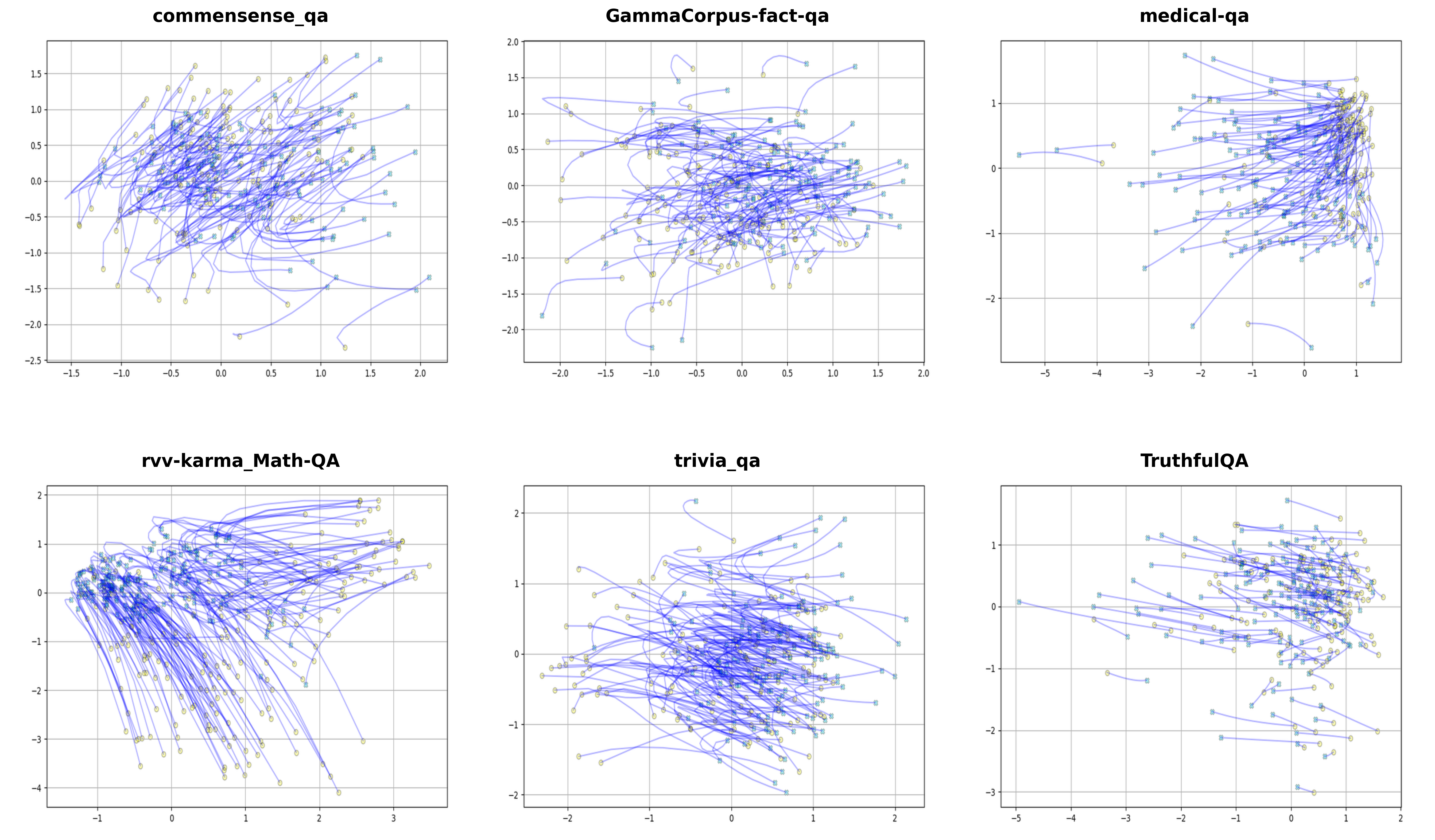}
    \caption{Input-to-Output Transformation Diagram Across Various QA Datasets (Yellow represents the starting point, and green represents the ending point)}
    \label{fig:combined_input_to_output_transformation_diagram}
\end{figure}

Principal Component Analysis (PCA) was used to reduce the dimensionality of the embedding vectors to two dimensions for visualization purposes. Figure~\ref{fig:combined_input_to_output_transformation_diagram} shows the PCA projections of embeddings from all six datasets, highlighting the clustering patterns and input-output transformation across diverse datasets.

\paragraph{Dataset Feature Analysis} 

The trajectory distributions for \texttt{commonsense\_qa} and \texttt{trivia\_qa} are relatively dispersed, indicating that the input-output mappings in commonsense reasoning tasks are complex, with diverse transformation paths. On the other hand, \texttt{medical-qa} shows a more concentrated endpoint distribution (with yellow points clustered on the right), suggesting that answers in the medical domain may have a more standardized format or structure. The \texttt{rvv-karma\_Math-QA} dataset exhibits a distinctly radial distribution, which likely reflects the logical reasoning path involved in solving mathematical problems. Lastly, the trajectories for \texttt{TruthfulQA} are relatively short and evenly distributed, suggesting that the transformation process for truthfulness verification is more straightforward.

\paragraph{Common Features} 

All datasets demonstrate a continuous transformation process from input (yellow points) to output (blue points), validating that Neural ODEs can effectively model the dynamic characteristics of LLMs. The trajectories generally exhibit nonlinear features, indicating that the input-output transformation in these QA tasks is complex.

\paragraph{Conclusion} 

In terms of modeling effectiveness, Neural ODE successfully captured the dynamic features of different QA tasks across various domains. The model was able to adapt to the specific characteristics of different datasets, demonstrating its versatility.

Regarding domain differences, the input-output transformations for different QA tasks revealed unique patterns. Specialized domains, such as medical and mathematical QA, showed more structured and regular transformation paths compared to commonsense reasoning tasks. This suggests that specialized fields benefit from more predictable and structured transformations, while more general domains, such as commonsense reasoning, involve more complex mappings.

\subsection{Experiment II: Results and analysis}

Experiment II examines the impact of Control Theory on the model's training process. We applied four control strategies, as detailed in the appendix, and compared the results with and without the use of control mechanisms. PCA was employed to visualize the output embeddings generated by the model in both scenarios, with and without the application of Control Theory. 

\paragraph{Trajectory Feature Analysis}

\begin{figure}[htbp]
    \centering
    \includegraphics[width=0.8\columnwidth]{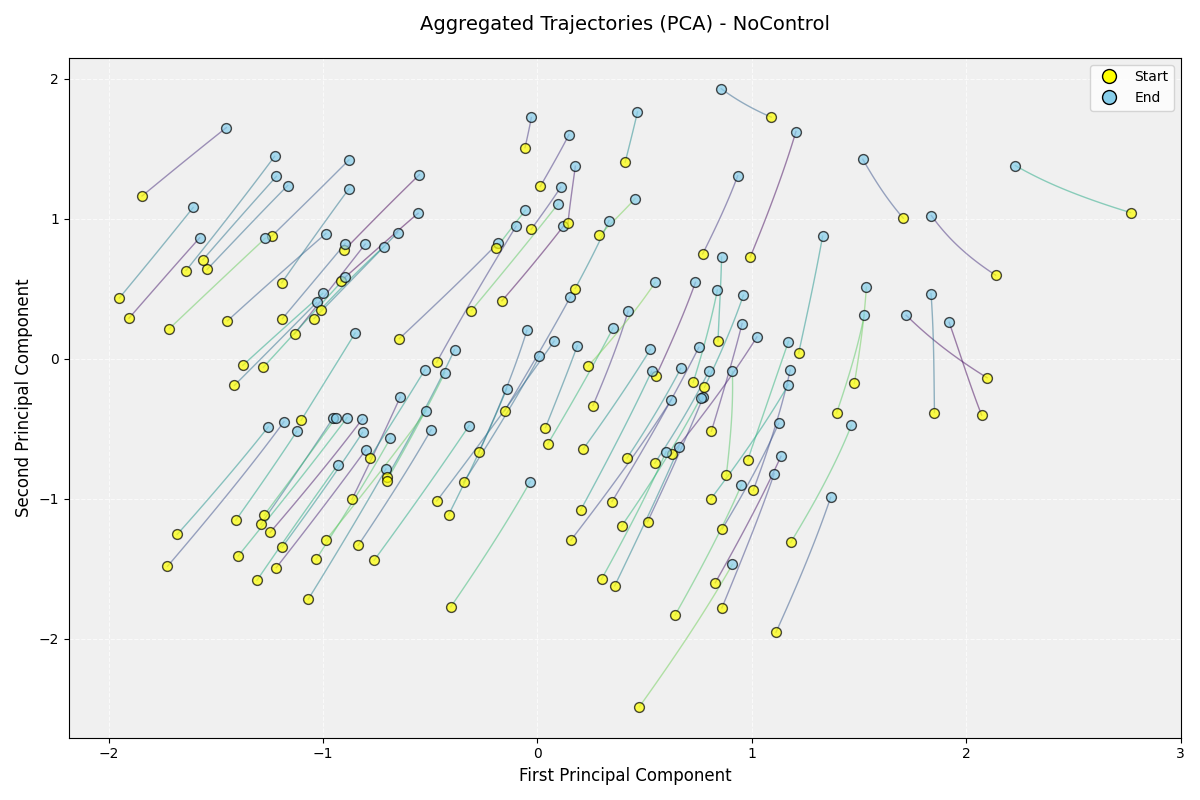}
    \caption{Trajectory Plots without Control}
    \label{fig:comparison_control}
\end{figure}

The analysis of the five trajectory plots—NoControl, LQRControl, MPCControl, RLControl, and SMControl—reveals several key patterns in the control mechanisms' performance. In the NoControl scenario, the trajectories appear scattered with no clear pattern, and the path from start to end is irregular. The absence of clear directionality and the significant variation in trajectory length highlight the instability of the model without control mechanisms.

\begin{figure}[htbp]
    \centering
    \includegraphics[width=0.8\columnwidth]{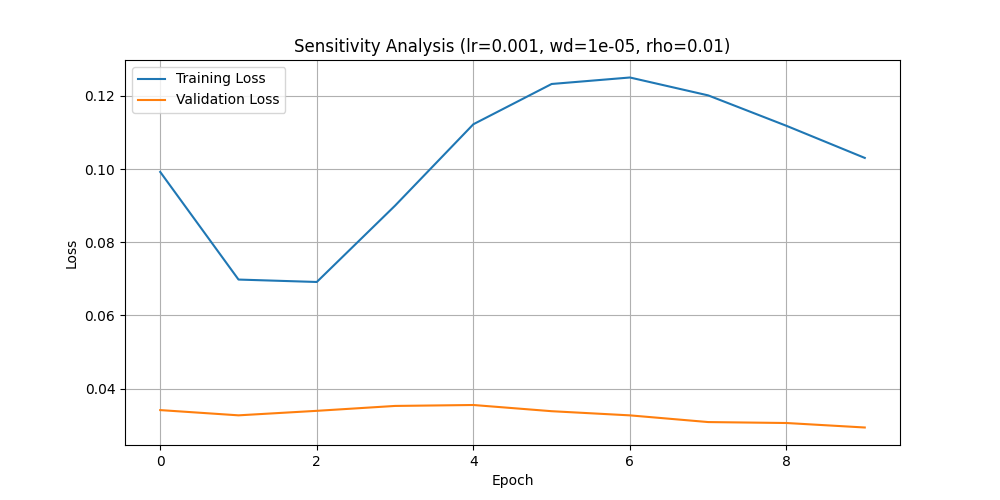}
    \caption{Trajectory Plots with LQR Control}
    \label{fig:comparison_control}
\end{figure}

When LQRControl is applied, the trajectories become notably smoother. The endpoint distribution is more concentrated, reflecting the symmetrical characteristics typical of linear control. This demonstrates that LQRControl provides more consistent results than NoControl.

\begin{figure}[htbp]
    \centering
    \includegraphics[width=0.8\columnwidth]{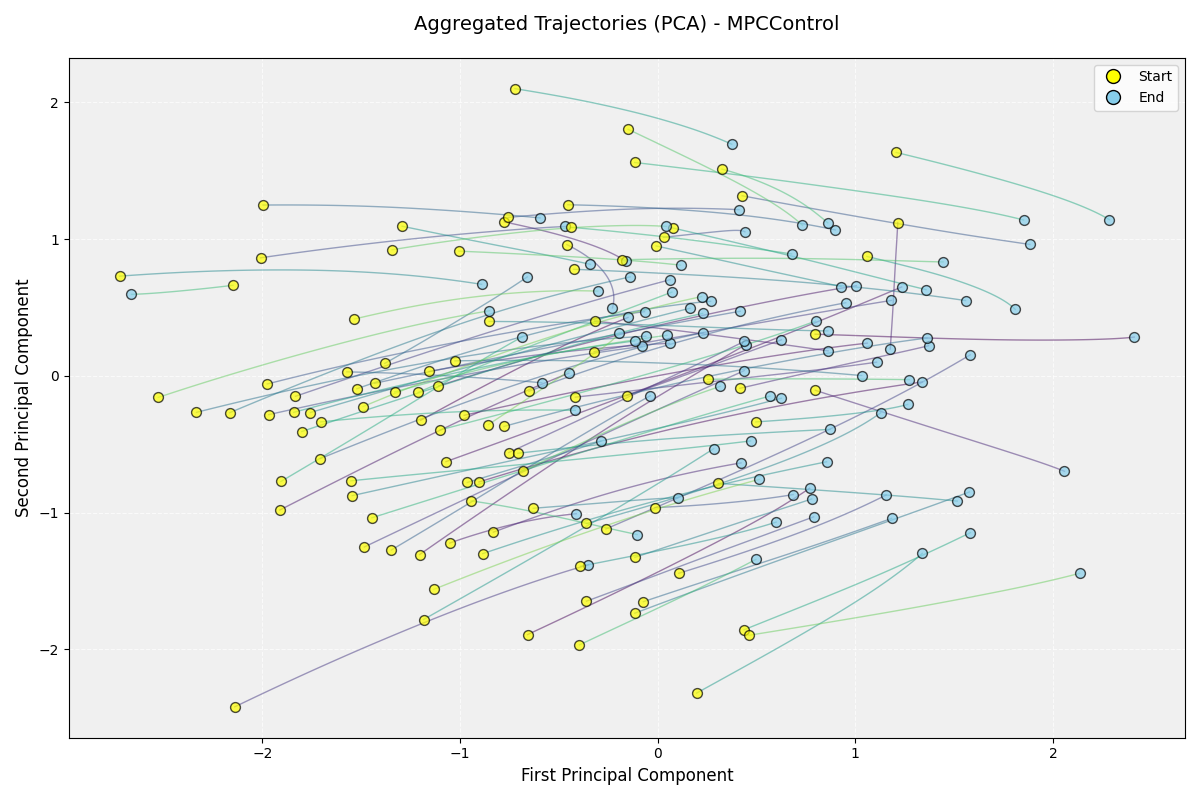}
    \caption{Trajectory Plots with MPC Control}
    \label{fig:comparison_control}
\end{figure}

With MPCControl, the trajectories exhibit a predictive feature, where the path planning becomes more rational, and the trajectory is more coherent. The endpoint distribution is moderate, indicating that MPCControl achieves a balanced optimization. The model demonstrates characteristics of model predictive control, where future trajectories are accounted for, guiding the model toward better output predictions.

\begin{figure}[htbp]
    \centering
    \includegraphics[width=0.8\columnwidth]{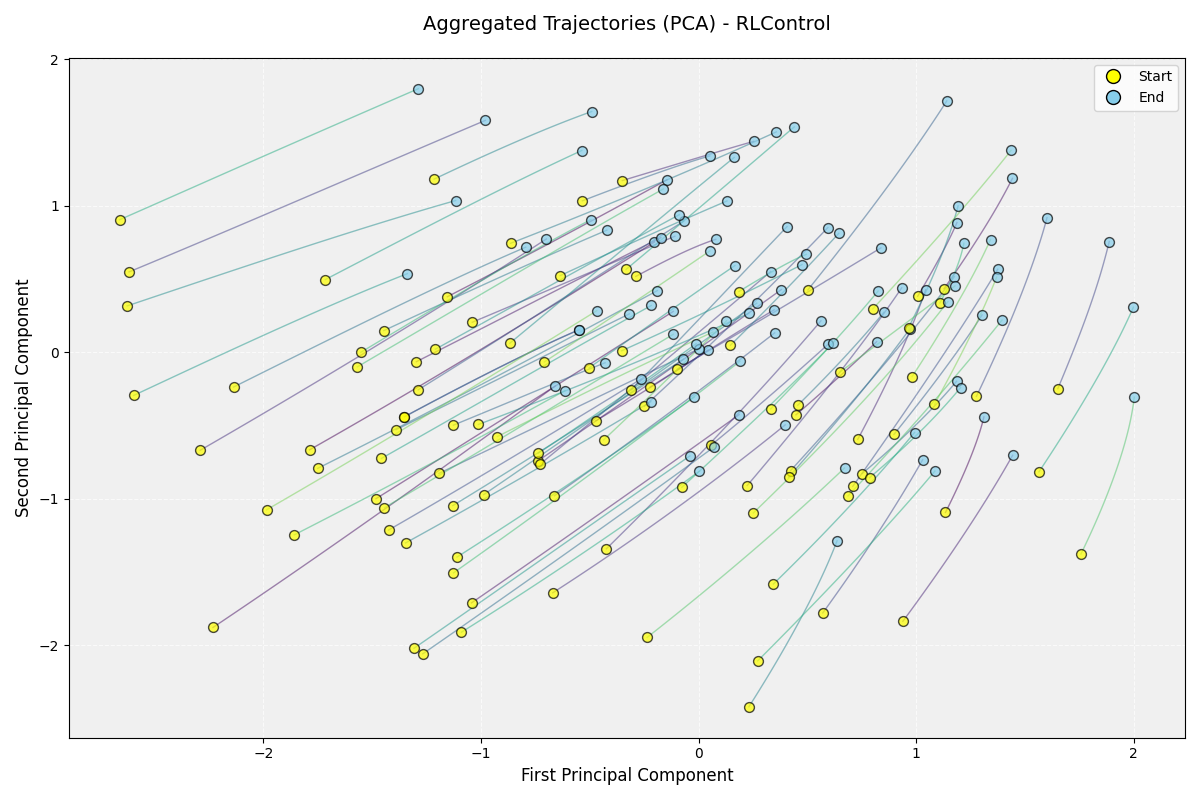}
    \caption{Trajectory Plots with RL Control}
    \label{fig:comparison_control}
\end{figure}

RLControl, in contrast, shows a stronger directionality. The model’s adaptation through reinforcement learning allows it to learn an effective control strategy, leading to a more concentrated endpoint distribution. The trajectories in this case reflect the self-adaptive nature of reinforcement learning, as the model learns and refines its control policy over time.

\begin{figure}[htbp]
    \centering
    \includegraphics[width=0.8\columnwidth]{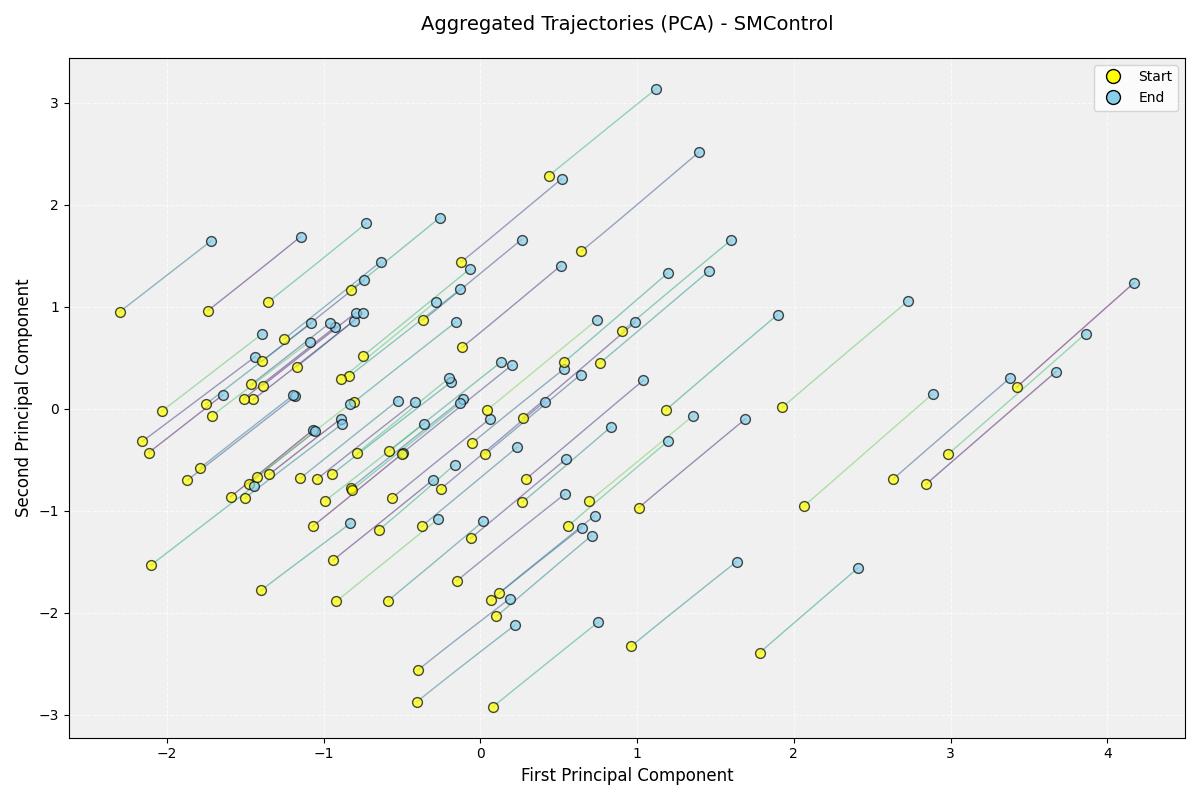}
    \caption{Trajectory Plots with SM Control}
    \label{fig:comparison_control}
\end{figure}

SMControl stands out as the most orderly and regular, with a clear sliding mode surface feature evident in the trajectories. The control effect is the most stable, and the directionality is the clearest, reflecting the strengths of sliding mode control in ensuring both stability and clarity in the model's output.

\paragraph{Conclusions}

In terms of overall evaluation, the application of control theory significantly enhances the model's output quality. Each control strategy exhibits its own unique advantages: SMControl excels in stability, RLControl in adaptability, MPCControl in prediction, and LQRControl in linear problems. These findings confirm that control theory plays a crucial role in enhancing both the reliability and interpretability of LLM outputs.

\section{Conclusion and Future Research}

This paper introduces a novel framework that integrates Neural Ordinary Differential Equations (Neural ODEs) with robust control theory to enhance the interpretability and reliability of Large Language Models (LLMs). Our empirical and theoretical analysis demonstrates that combining these approaches significantly improves model performance, stability, and adaptability. By implementing various control strategies, such as LQR, MPC, SM, and RL-based control, we validate their respective strengths in regulating LLM outputs, particularly in dynamic and complex environments.

Looking forward, future research can explore the development of hybrid control mechanisms, scalability optimizations for large-scale deployment, and the integration of advanced interpretability techniques. Additionally, addressing the computational challenges and enhancing generalization across diverse tasks will be crucial in refining this framework. By extending the applications to multimodal systems and real-time language processing, we can ensure that LLMs remain both reliable and interpretable, paving the way for safer and more effective AI systems in real-world applications.

\section{Limitation}
The proposed framework integrating Neural ODEs with control theory offers significant advancements in enhancing the interpretability and reliability of Large Language Models (LLMs). However, several limitations remain. First, the computational complexity of integrating continuous-time modeling with control mechanisms introduces substantial overhead, necessitating further optimization for large-scale models. Second, the task of parameter tuning for control strategies, such as feedback gains or prediction horizons, is challenging and requires further research to identify optimal settings across diverse tasks. Additionally, while the framework has shown promising results on specific datasets, its generalization across different LLM architectures and domains remains to be fully validated. Finally, although control theory improves model stability, further attention must be paid to ethical concerns such as fairness and accountability to ensure the responsible deployment of LLMs in high-stakes applications.

\section{Acknowledgements}
During the writing of this article, generative artificial intelligence tools were used to assist in language polishing and literature retrieval. The AI tool helped optimize the grammatical structure and expression fluency of limited paragraphs, and assisted in screening research literature in related fields. All AI-polished text content has been strictly reviewed by the author to ensure that it complies with academic standards and is accompanied by accurate citations. The core research ideas, method design and conclusion derivation of this article were independently completed by the author, and the AI tool did not participate in the proposal of any innovative research ideas or the creation of substantive content. The author is fully responsible for the academic rigor, data authenticity and citation integrity of the full text, and hereby declares that the generative AI tool is not a co-author of this study.

\bibliographystyle{ACM-Reference-Format}
\bibliography{sample-base}


\begin{thebibliography}{14}


\ifx \showCODEN    \undefined \def \showCODEN     #1{\unskip}     \fi
\ifx \showISBNx    \undefined \def \showISBNx     #1{\unskip}     \fi
\ifx \showISBNxiii \undefined \def \showISBNxiii  #1{\unskip}     \fi
\ifx \showISSN     \undefined \def \showISSN      #1{\unskip}     \fi
\ifx \showLCCN     \undefined \def \showLCCN      #1{\unskip}     \fi
\ifx \shownote     \undefined \def \shownote      #1{#1}          \fi
\ifx \showarticletitle \undefined \def \showarticletitle #1{#1}   \fi
\ifx \showURL      \undefined \def \showURL       {\relax}        \fi
\providecommand\bibfield[2]{#2}
\providecommand\bibinfo[2]{#2}
\providecommand\natexlab[1]{#1}
\providecommand\showeprint[2][]{arXiv:#2}

\bibitem[Beltagy et~al\mbox{.}(2020)]%
        {beltagy2020longformerlongdocumenttransformer}
\bibfield{author}{\bibinfo{person}{Iz Beltagy}, \bibinfo{person}{Matthew~E. Peters}, {and} \bibinfo{person}{Arman Cohan}.} \bibinfo{year}{2020}\natexlab{}.
\newblock \bibinfo{title}{Longformer: The Long-Document Transformer}.
\newblock
\showeprint[arxiv]{2004.05150}~[cs.CL]
\urldef\tempurl%
\url{https://arxiv.org/abs/2004.05150}
\showURL{%
\tempurl}


\bibitem[Brown et~al\mbox{.}(2020)]%
        {brown2020languagemodelsfewshotlearners}
\bibfield{author}{\bibinfo{person}{Tom~B. Brown}, \bibinfo{person}{Benjamin Mann}, \bibinfo{person}{Nick Ryder}, \bibinfo{person}{Melanie Subbiah}, \bibinfo{person}{Jared Kaplan}, \bibinfo{person}{Prafulla Dhariwal}, \bibinfo{person}{Arvind Neelakantan}, \bibinfo{person}{Pranav Shyam}, \bibinfo{person}{Girish Sastry}, \bibinfo{person}{Amanda Askell}, \bibinfo{person}{Sandhini Agarwal}, \bibinfo{person}{Ariel Herbert-Voss}, \bibinfo{person}{Gretchen Krueger}, \bibinfo{person}{Tom Henighan}, \bibinfo{person}{Rewon Child}, \bibinfo{person}{Aditya Ramesh}, \bibinfo{person}{Daniel~M. Ziegler}, \bibinfo{person}{Jeffrey Wu}, \bibinfo{person}{Clemens Winter}, \bibinfo{person}{Christopher Hesse}, \bibinfo{person}{Mark Chen}, \bibinfo{person}{Eric Sigler}, \bibinfo{person}{Mateusz Litwin}, \bibinfo{person}{Scott Gray}, \bibinfo{person}{Benjamin Chess}, \bibinfo{person}{Jack Clark}, \bibinfo{person}{Christopher Berner}, \bibinfo{person}{Sam McCandlish}, \bibinfo{person}{Alec Radford}, \bibinfo{person}{Ilya Sutskever},
  {and} \bibinfo{person}{Dario Amodei}.} \bibinfo{year}{2020}\natexlab{}.
\newblock \bibinfo{title}{Language Models are Few-Shot Learners}.
\newblock
\showeprint[arxiv]{2005.14165}~[cs.CL]
\urldef\tempurl%
\url{https://arxiv.org/abs/2005.14165}
\showURL{%
\tempurl}


\bibitem[Chen et~al\mbox{.}(2019)]%
        {chen2019neuralordinarydifferentialequations}
\bibfield{author}{\bibinfo{person}{Ricky T.~Q. Chen}, \bibinfo{person}{Yulia Rubanova}, \bibinfo{person}{Jesse Bettencourt}, {and} \bibinfo{person}{David Duvenaud}.} \bibinfo{year}{2019}\natexlab{}.
\newblock \bibinfo{title}{Neural Ordinary Differential Equations}.
\newblock
\showeprint[arxiv]{1806.07366}~[cs.LG]
\urldef\tempurl%
\url{https://arxiv.org/abs/1806.07366}
\showURL{%
\tempurl}


\bibitem[Hashimoto et~al\mbox{.}(2024)]%
        {hashimoto2024unificationsymmetriesinsideneural}
\bibfield{author}{\bibinfo{person}{Koji Hashimoto}, \bibinfo{person}{Yuji Hirono}, {and} \bibinfo{person}{Akiyoshi Sannai}.} \bibinfo{year}{2024}\natexlab{}.
\newblock \bibinfo{title}{Unification of Symmetries Inside Neural Networks: Transformer, Feedforward and Neural ODE}.
\newblock
\showeprint[arxiv]{2402.02362}~[cs.LG]
\urldef\tempurl%
\url{https://arxiv.org/abs/2402.02362}
\showURL{%
\tempurl}


\bibitem[Kevian et~al\mbox{.}(2024)]%
        {kevian2024capabilitieslargelanguagemodels}
\bibfield{author}{\bibinfo{person}{Darioush Kevian}, \bibinfo{person}{Usman Syed}, \bibinfo{person}{Xingang Guo}, \bibinfo{person}{Aaron Havens}, \bibinfo{person}{Geir Dullerud}, \bibinfo{person}{Peter Seiler}, \bibinfo{person}{Lianhui Qin}, {and} \bibinfo{person}{Bin Hu}.} \bibinfo{year}{2024}\natexlab{}.
\newblock \bibinfo{title}{Capabilities of Large Language Models in Control Engineering: A Benchmark Study on GPT-4, Claude 3 Opus, and Gemini 1.0 Ultra}.
\newblock
\showeprint[arxiv]{2404.03647}~[math.OC]
\urldef\tempurl%
\url{https://arxiv.org/abs/2404.03647}
\showURL{%
\tempurl}


\bibitem[Liang et~al\mbox{.}(2024)]%
        {liang2024controllabletextgenerationlarge}
\bibfield{author}{\bibinfo{person}{Xun Liang}, \bibinfo{person}{Hanyu Wang}, \bibinfo{person}{Yezhaohui Wang}, \bibinfo{person}{Shichao Song}, \bibinfo{person}{Jiawei Yang}, \bibinfo{person}{Simin Niu}, \bibinfo{person}{Jie Hu}, \bibinfo{person}{Dan Liu}, \bibinfo{person}{Shunyu Yao}, \bibinfo{person}{Feiyu Xiong}, {and} \bibinfo{person}{Zhiyu Li}.} \bibinfo{year}{2024}\natexlab{}.
\newblock \bibinfo{title}{Controllable Text Generation for Large Language Models: A Survey}.
\newblock
\showeprint[arxiv]{2408.12599}~[cs.CL]
\urldef\tempurl%
\url{https://arxiv.org/abs/2408.12599}
\showURL{%
\tempurl}


\bibitem[Lipton(2017)]%
        {lipton2017mythosmodelinterpretability}
\bibfield{author}{\bibinfo{person}{Zachary~C. Lipton}.} \bibinfo{year}{2017}\natexlab{}.
\newblock \bibinfo{title}{The Mythos of Model Interpretability}.
\newblock
\showeprint[arxiv]{1606.03490}~[cs.LG]
\urldef\tempurl%
\url{https://arxiv.org/abs/1606.03490}
\showURL{%
\tempurl}


\bibitem[Lundberg and Lee(2017)]%
        {lundberg2017unifiedapproachinterpretingmodel}
\bibfield{author}{\bibinfo{person}{Scott Lundberg} {and} \bibinfo{person}{Su-In Lee}.} \bibinfo{year}{2017}\natexlab{}.
\newblock \bibinfo{title}{A Unified Approach to Interpreting Model Predictions}.
\newblock
\showeprint[arxiv]{1705.07874}~[cs.AI]
\urldef\tempurl%
\url{https://arxiv.org/abs/1705.07874}
\showURL{%
\tempurl}


\bibitem[Meng et~al\mbox{.}(2023)]%
        {meng2023locatingeditingfactualassociations}
\bibfield{author}{\bibinfo{person}{Kevin Meng}, \bibinfo{person}{David Bau}, \bibinfo{person}{Alex Andonian}, {and} \bibinfo{person}{Yonatan Belinkov}.} \bibinfo{year}{2023}\natexlab{}.
\newblock \bibinfo{title}{Locating and Editing Factual Associations in GPT}.
\newblock
\showeprint[arxiv]{2202.05262}~[cs.CL]
\urldef\tempurl%
\url{https://arxiv.org/abs/2202.05262}
\showURL{%
\tempurl}


\bibitem[Rubanova et~al\mbox{.}(2019)]%
        {rubanova2019latentodesirregularlysampledtime}
\bibfield{author}{\bibinfo{person}{Yulia Rubanova}, \bibinfo{person}{Ricky T.~Q. Chen}, {and} \bibinfo{person}{David Duvenaud}.} \bibinfo{year}{2019}\natexlab{}.
\newblock \bibinfo{title}{Latent ODEs for Irregularly-Sampled Time Series}.
\newblock
\showeprint[arxiv]{1907.03907}~[cs.LG]
\urldef\tempurl%
\url{https://arxiv.org/abs/1907.03907}
\showURL{%
\tempurl}


\bibitem[Sundararajan et~al\mbox{.}(2017)]%
        {sundararajan2017axiomaticattributiondeepnetworks}
\bibfield{author}{\bibinfo{person}{Mukund Sundararajan}, \bibinfo{person}{Ankur Taly}, {and} \bibinfo{person}{Qiqi Yan}.} \bibinfo{year}{2017}\natexlab{}.
\newblock \bibinfo{title}{Axiomatic Attribution for Deep Networks}.
\newblock
\showeprint[arxiv]{1703.01365}~[cs.LG]
\urldef\tempurl%
\url{https://arxiv.org/abs/1703.01365}
\showURL{%
\tempurl}


\bibitem[Tenney et~al\mbox{.}(2019)]%
        {tenney2019learncontextprobingsentence}
\bibfield{author}{\bibinfo{person}{Ian Tenney}, \bibinfo{person}{Patrick Xia}, \bibinfo{person}{Berlin Chen}, \bibinfo{person}{Alex Wang}, \bibinfo{person}{Adam Poliak}, \bibinfo{person}{R~Thomas McCoy}, \bibinfo{person}{Najoung Kim}, \bibinfo{person}{Benjamin~Van Durme}, \bibinfo{person}{Samuel~R. Bowman}, \bibinfo{person}{Dipanjan Das}, {and} \bibinfo{person}{Ellie Pavlick}.} \bibinfo{year}{2019}\natexlab{}.
\newblock \bibinfo{title}{What do you learn from context? Probing for sentence structure in contextualized word representations}.
\newblock
\showeprint[arxiv]{1905.06316}~[cs.CL]
\urldef\tempurl%
\url{https://arxiv.org/abs/1905.06316}
\showURL{%
\tempurl}


\bibitem[Vig(2019)]%
        {vig2019visualizingattentiontransformerbasedlanguage}
\bibfield{author}{\bibinfo{person}{Jesse Vig}.} \bibinfo{year}{2019}\natexlab{}.
\newblock \bibinfo{title}{Visualizing Attention in Transformer-Based Language Representation Models}.
\newblock
\showeprint[arxiv]{1904.02679}~[cs.HC]
\urldef\tempurl%
\url{https://arxiv.org/abs/1904.02679}
\showURL{%
\tempurl}


\bibitem[Wang et~al\mbox{.}(2024)]%
        {wang2024learningadaptivehydrodynamicmodels}
\bibfield{author}{\bibinfo{person}{Cong Wang}, \bibinfo{person}{Aoming Liang}, \bibinfo{person}{Fei Han}, \bibinfo{person}{Xinyu Zeng}, \bibinfo{person}{Zhibin Li}, \bibinfo{person}{Dixia Fan}, {and} \bibinfo{person}{Jens Kober}.} \bibinfo{year}{2024}\natexlab{}.
\newblock \bibinfo{title}{Learning Adaptive Hydrodynamic Models Using Neural ODEs in Complex Conditions}.
\newblock
\showeprint[arxiv]{2410.00490}~[cs.RO]
\urldef\tempurl%
\url{https://arxiv.org/abs/2410.00490}
\showURL{%
\tempurl}


\end{thebibliography}

\appendix

\section{Detailed Experiment}

\begin{table}[htbp]
    \centering
    \caption{Overview of QA Datasets Used in Experiment I}
    \label{tab:experiment1_datasets}
    \begin{tabularx}{\linewidth}{lX}
        \toprule
        \textbf{Dataset Name} & \textbf{Domain}  \\
        \midrule
        commonsense\_qa & Commonsense Reasoning \\
        GammaCorpus-fact-qa & General Knowledge, Fact-Checking \\
        medical-qa & Medical, Healthcare \\
        rvv-karma\_Math-QA & Mathematics, Logical Reasoning \\
        trivia\_qa & Trivia, Common Knowledge \\
        TruthfulQA & Truthfulness, Ethics \\
        \bottomrule
    \end{tabularx}
\end{table}

\section{Appendix:Control Strategies Summary}

In this section, we present four control strategies implemented for regulating the outputs of Large Language Models (LLMs): Linear Quadratic Regulator (LQR), Model Predictive Control (MPC), Sliding Mode Control (SMC), and Reinforcement Learning (RL) based control. Each of these methods employs different approaches to improve the model's performance, stability, and adaptability.

\subsection{Linear Quadratic Regulator Control (LQRControl)}
Linear Quadratic Regulator (LQR) is an optimal control strategy that minimizes a quadratic cost function to stabilize the system. The objective is to penalize both the state deviations and the control effort. The control input \( u \) is calculated as the negative feedback of the error between the desired state \( q \) and the model output \( y \), scaled by the feedback gain matrix \( K \).

The LQRControl class has three primary parameters:
- \( \mathbf{Q} \in \mathbb{R}^{n \times n} \): A state penalty matrix that penalizes deviations of the system state from the desired state.
- \( \mathbf{R} \in \mathbb{R}^{m \times m} \): A control effort penalty matrix that penalizes large control actions.
- \( \mathbf{K} \in \mathbb{R}^{n \times m} \): A feedback gain matrix that defines how much influence the control input has over the system's state.

The control law is defined as follows:
\[
u = - \mathbf{K} \cdot (q - y) \cdot \sigma \left( \sum \mathbf{Q} \right) / \left( \sum \mathbf{R} + \epsilon \right)
\]
where \( \sigma \) is the sigmoid activation function, and \( \epsilon \) is a small constant to avoid division by zero.

LQRControl is effective in stabilizing systems where the relationship between the state and control is linear, and it is particularly useful in scenarios that require precise error correction.

\subsection{Model Predictive Control (MPCControl)}
Model Predictive Control (MPC) uses a model of the system to predict future states and optimize the control input over a finite time horizon. MPC computes the control input by solving an optimization problem that minimizes a cost function over the predicted trajectory. This method is highly effective in systems where future behavior can be predicted and adjusted.

The MPCControl class uses the following parameters:
- \( \mathbf{Horizon} \in \mathbb{N} \): The prediction horizon over which the system's behavior is forecasted.
- \( \mathbf{State Predictor} \): A neural network used to predict future system states based on the current state and past control inputs.
- \( \mathbf{Controller} \): A neural network that computes the optimal control action by considering the predicted states.

The MPC control input is computed as:
\[
u = \arg\min_u \sum_{t=1}^{\text{Horizon}} \left( \|y_t - q_t\|^2 + \lambda_u \|u_t\|^2 \right)
\]
where \( y_t \) is the predicted state at time step \( t \), \( q_t \) is the desired state, and \( \lambda_u \) is a regularization parameter.

MPCControl is particularly suitable for systems that require optimization over a planning horizon, such as robotics and autonomous systems.

\subsection{Sliding Mode Control (SMControl)}
Sliding Mode Control (SMC) is a robust control technique designed to handle systems with uncertainties or disturbances. SMC forces the system state to "slide" along a predefined sliding surface, ensuring robustness and stability. The control input is determined based on the system's error and the sliding surface function, which enforces the desired behavior.

The SMControl class involves the following parameters:
- \( \mathbf{\lambda} \in \mathbb{R}^n \): The scaling factor for the sliding surface, which governs the control input's sensitivity.
- \( \mathbf{\eta} \in \mathbb{R}^n \): A parameter that adjusts the system's response to the error.
- \( \mathbf{\phi} \in \mathbb{R}^n \): A parameter that controls the non-linearity in the sliding surface.

The control law is defined as:
\[
u = \lambda \cdot s + k \cdot \text{sat}(s / \phi)
\]
where \( s = q - y \) is the error, and \( \text{sat}(\cdot) \) is the saturation function, typically \( \text{sat}(x) = \tanh(x) \).

SMControl is effective for systems requiring high robustness against uncertainties and disturbances, such as in automotive or aerospace applications.

\subsection{Reinforcement Learning Control (RLControl)}
Reinforcement Learning (RL) Control leverages reinforcement learning to adaptively learn the optimal control policy based on feedback from the environment. The RL controller utilizes a value network to estimate the expected future reward and a policy network to decide the control actions. The advantage function adjusts the control input by calculating the difference between expected and actual outcomes.

The RLControl class includes the following components:
- \( \mathbf{\gamma} \in [0, 1] \): The discount factor that determines the importance of future rewards.
- \( \mathbf{Value Network} \): A neural network that estimates the value of the current state.
- \( \mathbf{Policy Network} \): A neural network that generates the control action based on the current state and the desired state.

The control input is computed as:
\[
u = \text{policy}(y, q) \cdot \sigma(\| q - y \| - \text{value}(y))
\]
where \( \sigma \) is the sigmoid activation function, \( \| q - y \| \) is the norm of the error, and \( \text{value}(y) \) is the output of the value network.

RLControl is suitable for complex, dynamic environments where the model must continuously learn and adapt to new situations.

\subsection{Conclusion}
Each control strategy offers distinct advantages depending on the application scenario. LQRControl is well-suited for linear systems with simple control tasks, while MPCControl provides superior performance in systems requiring trajectory optimization and prediction. SMControl excels in environments that demand robustness and stability, and RLControl is ideal for tasks that require adaptive learning and optimization in complex, dynamic settings. The integration of these control strategies demonstrates substantial improvements in model performance, stability, and interpretability across a variety of tasks.

\end{document}